\documentclass[runningheads]{llncs}
\usepackage{epsf}
\usepackage{multirow}
\usepackage{amssymb}
\usepackage{graphicx}

\usepackage{url}

\begin{document}

\title{WiSeBE:~Window-based Sentence~Boundary~Evaluation}

\author{Carlos-Emiliano Gonz\'alez-Gallardo\inst{1,2} \and
Juan-Manuel Torres-Moreno\inst{1,2}}
\authorrunning{C.E. Gonz\'alez-Gallardo and J.M. Torres-Moreno}
%
\institute{LIA - Universit\'e d'Avignon et des Pays de Vaucluse, 339 chemin des Meinajaries, 84140, Avignon, France\\
\email{carlos-emiliano.gonzalez-gallardo@alumni.univ-avignon.fr\\
juan-manuel.torres@univ-avignon.fr} \and
D\'epartement de GIGL, \'Ecole Polytechnique de Montr\'eal,\\
C.P. 6079, succ. Centre-ville, Montr\'eal (Qu\'ebec) H3C 3A7 Canada}

\maketitle

\begin{abstract}

Sentence Boundary Detection (SBD) has been a major research topic since Automatic Speech Recognition transcripts have been used for further Natural Language Processing tasks like Part of Speech Tagging, Question Answering or Automatic Summarization.
But what about evaluation?
Do standard evaluation metrics like precision, recall,  F-score or classification error; and more important, evaluating an automatic system against a unique reference is enough to conclude how well a SBD system is performing given the final application of the transcript?
In this paper we propose Window-based Sentence Boundary Evaluation (WiSeBE), a semi-supervised metric for evaluating Sentence Boundary Detection systems based on multi-reference (dis)agreement.
We evaluate and compare the performance of different SBD systems over a set of Youtube transcripts using WiSeBE and standard metrics.
This double evaluation gives an understanding of how WiSeBE is a more reliable metric for the SBD task.


\end{abstract}

\section{Introduction}
\label{sec:Introduction}

The goal of Automatic Speech Recognition (ASR) is to transform spoken data into a written representation, thus enabling natural human-machine interaction \cite{yu2016automatic} with further Natural Language Processing (NLP) tasks.
Machine translation, question answering, semantic parsing, POS tagging, sentiment analysis and automatic text summarization; originally developed to work with formal written texts, can be applied over the transcripts made by ASR systems \cite{brum2016sentiment,stevenson2000experiments,wang2010automatic}.
However, before applying any of these NLP tasks a segmentation process called Sentence Boundary Detection (SBD) should be performed over ASR transcripts to reach a minimal syntactic information in the text.

To measure the performance of a SBD system, the automatically segmented transcript is evaluated against a single reference normally done by a human.
But given a transcript, does it exist a unique reference?
Or, is it possible that the same transcript could be segmented in five different ways by five different people in the same conditions?
If so, which one is correct; and more important, how to fairly evaluate the automatically segmented transcript? These questions are the foundations of Window-based Sentence Boundary Evaluation (WiSeBE), a new semi-supervised metric for evaluating SBD systems based on multi-reference (dis)agreement.

The rest of this article is organized as follows.
In Section \ref{sec:SBD} we set the frame of SBD and how it is normally evaluated.
WiSeBE is formally described in Section \ref{sec:WiSeBE}, followed by a multi-reference evaluation in Section \ref{sec:exp}.
Further analysis of WiSeBE and discussion over the method and alternative multi-reference evaluation is presented in Section \ref{sec:disc}.
Finally, Section \ref{sec:conc} concludes the paper.

\section{Sentence Boundary Detection}
\label{sec:SBD}

Sentence Boundary Detection (SBD) has been a major research topic science ASR moved to more general domains as conversational speech \cite{meteer1996modeling,shriberg1996word,stolcke1996automatic}.
Performance of ASR systems has improved over the years with the inclusion and combination of new Deep Neural Networks methods \cite{fohr2017new,hinton2012deep,yu2016automatic}.
As a general rule, the output of ASR systems lacks of any syntactic information such as capitalization and sentence boundaries, showing the interst of ASR systems to obtain the correct sequence of words with almost no concern of the overall structure of the document \cite{gotoh2000sentence}.

Similar to SBD is the Punctuation Marks Disambiguation (PMD) or Sentence Boundary Disambiguation.
This task aims to segment a formal written text into well formed sentences based on the existent punctuation marks \cite{kiss2006unsupervised,palmer1994adaptive,palmer1997adaptive,treviso2017evaluating}. 
In this context a sentence is defined (for English) by the Cambridge Dictionary\footnote{https://dictionary.cambridge.org/} as:

\begin{quote} 
\centering 
    \textit{``a group of words, usually containing a verb, that expresses a thought in the form of a statement, question, instruction, or exclamation and starts with a capital letter when written''}.
\end{quote}

PMD carries certain complications, some given the ambiguity of punctuation marks within a sentence. A period can denote an acronym, an abbreviation, the end of the sentence or a combination of them as in the following example:

\begin{quote} 
\centering 
    \textit{The U.S. president, Mr. Donald Trump, is meeting with the F.B.I. director Christopher A. Wray next Thursday at 8p.m.}
\end{quote}

However its difficulties, DPM profits of morphological and lexical information to achieve a correct sentence segmentation.
By contrast, segmenting an ASR transcript should be done without any (or almost any) lexical information and a flurry definition of sentence.

The obvious division in spoken language may be considered speaker utterances. However, in a normal conversation or even in a monologue, the way ideas are organized differs largely from written text. This differences, added to disfluencies like revisions, repetitions, restarts, interruptions and hesitations make the definition of a sentence unclear thus complicating the segmentation task \cite{strassel2003simple}.
Table \ref{tb:sbdexample} exemplifies some of the difficulties that are present when working with spoken language.

\begin{table}
\caption{Sentnce Boundary Detection example}
\label{tb:sbdexample}
\begin{center}
\begin{tabular}{p{6cm}|p{6cm}}
\hline
Speech transcript & SBD applied to transcript \\
\hline
two two women can look out after a kid so bad as a man and a woman can so you can have a you can have a mother and a father that that still don't do right with the kid and you can have to men that can so as long as the love each other as long as they love each other it doesn't matter &
two // two women can look out after a kid so bad as a man and a woman can //
so you can have a // you can have a mother and a father that // that still don't do right with the kid and you can have to men that can // so as long as the love each other // as long as they love each other it doesn't matter // \\
\hline
\end{tabular}

\end{center}
\end{table}

Stolcke \& Shriberg \cite{stolcke1996automatic} considered a set of linguistic structures as segments including the following list:

\begin{itemize}
    \item Complete sentences
    \item Stand-alone sentences
    \item Disfluent sentences aborted in mid-utterance
    \item Interjections
    \item Back-channel responses
\end{itemize}

In \cite{meteer1996modeling},  Meteer \& Iyer divided speaker utterances into segments, consisting each of a single independent clause. A segment was considered to begin either at the beginning of an utterance, or after the end of the preceding segment. Any dysfluency between the end of the previous segments and the begging of current one was considered part of the current segments.

Rott \& \v Cerva \cite{rott2016speech} aimed to summarize news delivered orally segmenting the transcripts into \textit{``something that is similar to sentences''}.
They used a syntatic analyzer to identify the phrases within the text.

A wide study focused in unbalanced data for the SBD task was performed by Liu \textit{et al.} \cite{liu2006study}.
During this study they followed the segmentation scheme proposed by the Linguistic Data Consortium\footnote{https://www.ldc.upenn.edu/} on the Simple Metadata Annotation Specification V5.0 guideline (SimpleMDE\_V5.0) \cite{strassel2003simple}, dividing the transcripts in Semantic Units.

A Semantic Unit (SU) is considered to be an atomic element of the transcript that manages to express a complete thought or idea on the part of the speaker \cite{strassel2003simple}.
Sometimes a SU corresponds to the equivalent of a sentence in written text, but other times (the most part of them) a SU corresponds to a phrase or a single word.

SUs seem to be an inclusive conception of a segment, they embrace different previous segment definitions and are flexible enough to deal with the majority of spoken language troubles.
For these reasons we will adopt SUs as our segment definition.

\subsection{Sentence Boundary Evaluation}

SBD research has been focused on two different aspects; features and methods.
Regarding the features, some work focused on acoustic elements like pauses duration, fundamental frequencies, energy, rate of speech, volume change and speaker turn \cite{jamil2015sentence,klejch2016punctuated,kolavr2004automatic}. 

The other kind of features used in SBD are textual or lexical features.
They rely on the transcript content to extract features like bag-of-word, POS tags or word embeddings \cite{gonzalez2018sentence,klejch2016punctuated,lu2010better,mrozinski2006automatic,rott2016speech,stolcke1996automatic,ueffing2013improved}.
Mixture of acoustic and lexical features have also been explored \cite{bohac2012post,kolavr2012development,kolavr2004automatic,xu2014deep}, which is advantageous when both audio signal and transcript are available.

With respect to the methods used for SBD, they mostly rely on statistical/neural machine translation \cite{klejch2016punctuated,peitz2014better}, language models \cite{gotoh2000sentence,liu2006study,mrozinski2006automatic,stolcke1996automatic}, conditional random fields \cite{lu2010better,nicola2013improved,ueffing2013improved} and deep neural networks \cite{che2016punctuation,gonzalez2018sentence,treviso2017evaluating}.

Despite their differences in features and/or methodology, almost all previous cited research share a common element; the evaluation methodology. Metrics as Precision, Recall, F1-score, Classification Error Rate and Slot Error Rate (SER) are used to evaluate the proposed system against one reference.
As discussed in Section \ref{sec:Introduction}, further NLP tasks rely on the result of SBD, meaning that is crucial to have a good segmentation.
But comparing the output of a system against a unique reference will provide a reliable score to decide if the system is good or bad? 

Bohac \textit{et al.} \cite{bohac2012post} compared the human ability to punctuate recognized spontaneous speech.
They asked 10 people (correctors) to punctuate about 30 minutes of ASR transcripts in Czech.
For an average of 3,962 words, the punctuation marks placed by correctors varied between 557 and 801; this means a difference of 244 segments for the same transcript.
Over all correctors, the absolute consensus for period (.) was only 4.6\% caused by the replacement of other punctuation marks as semicolons (;) and exclamation marks (!).
These results are understandable if we consider the difficulties presented previously in this section.


To our knowledge, the amount of studies that have tried to target the sentence boundary evaluation with a multi-reference approach is very small.
In \cite{bohac2012post}, Bohac \textit{et al.} evaluated the overall punctuation accuracy for Czech in a straightforward multi-reference framework.
They considered a period (.) valid if at least five of their 10 correctors agreed on its position.

Kol\'a\v r \& Lamel \cite{kolavr2012development} considered two independent references to evaluate their system and proposed two approaches.
The fist one was to calculate the SER for each of one the two available references and then compute their mean.
They found this approach to be very strict because for those boundaries where no agreement between references existed, the system was going to be partially wrong even the fact that it has correctly predicted the boundary.
Their second approach tried to moderate the number of unjust penalizations.
For this case, a classification was considered incorrect only if it didn't match either of the two references.

These two examples exemplify the real need and some straightforward solutions for multi-reference evaluation metrics.
However, we think that it is possible to consider in a more inclusive approach the similarities and differences that multiple references could provide into a sentence boundary evaluation protocol.

\section{Window-Based Sentence Boundary Evaluation}
\label{sec:WiSeBE}

Window-Based Sentence Boundary Evaluation (WiSeBE) is a semi-automatic multi-reference sentence boundary evaluation protocol which considers the performance of a candidate segmentation over a set of segmentation references and the agreement between those references.

Let $\textbf{R} = \{R_1, R_2, ..., R_m\}$ be the set of all available references given a transcript $T = \{t_1, t_2, ..., t_n\}$, where $t_j$ is the $j^{th}$ word in the transcript; a reference $R_i$ is defined as a binary vector in terms of the existent SU boundaries in $T$.

\begin{equation}
    R_i = \{b_1, b_2, ..., b_n\}
\end{equation}

where
\[
    b_j=\left\{
        \begin{array}{ll}
        1 & \textrm{if }t_j\textrm{ is a boundary}\\ 
        0 & \textrm{otherwise} 
        \end{array}
    \right.
\]

\noindent Given a transcript $T$, the candidate segmentation  $C_T$ is defined similar to $R_i$.

\begin{equation}
    C_T = \{b_1, b_2, ..., b_n\}
\end{equation}

where
\[
    b_j=\left\{
        \begin{array}{ll}
        1 & \textrm{if }t_j\textrm{ is a boundary}\\ 
        0 & \textrm{otherwise} 
        \end{array}
    \right.
\]

\subsection{General Reference and Agreement Ratio}

A General Reference ($R_G$) is then constructed to calculate the agreement ratio between all references in.
It is defined by the boundary frequencies of each reference $R_i \in \textbf{R}$.

\begin{equation}
    R_G = \{d_1, d_2, ..., d_n\}
   \label{eq:RG}
\end{equation}

where

\begin{equation}
    d_j = \sum_{i=1}^{m}t_{ij}  \quad \forall t_j \in T, \quad   d_j =  [0,m] \;
\end{equation}

\noindent The Agreement Ratio ($R_{G_{AR}}$) is needed to get a numerical value of the distribution of SU boundaries over $\textbf{R}$. A value of $R_{G_{AR}}$ close to $0$ means a low agreement between references in $\textbf{R}$, while $R_{G_{AR}} = 1$ means a perfect agreement ($\forall R_i \in \textbf{R}, R_i=R_{i+1} | i=1, ..., m-1$) in $\textbf{R}$.

\begin{equation}
    \label{eq:AR}
    R_{G_{AR}} = \frac{R_{G_{PB}}}{R_{G_{HA}}} \;
\end{equation}

\noindent In the equation above, $R_{G_{PB}}$ corresponds to the ponderated common boundaries of $R_G$ and $R_{G_{HA}}$ to its hypothetical maximum agreement.

\begin{equation}
    R_{G_{PB}} = \sum_{j=1}^{n} d_j \left [ d_j \geq 2 \right ] \;
\end{equation}

\begin{equation}
    R_{G_{HA}} = m \times \sum_{d_j \in R_G}1\left [  d_j \neq  0 \right ] \;
\end{equation}

\subsection{Window-Boundaries Reference}

In Section \ref{sec:SBD} we discussed about how disfluencies complicate SU segmentation.
In a multi-reference environment this causes disagreement between references around a same SU boundary.
The way WiSeBE handle disagreements produced by disfluencies is with a Window-boundaries Reference ($R_W$) defined as:

\begin{equation}
    R_W = \{w_1, w_2, ..., w_p\}
   \label{eq:RW}
\end{equation}

where each window $w_k$ considers one or more boundaries $d_j$ from $R_G$ with a window separation limit equal to $R_{W_l}$.

\begin{equation}
    w_k = \{d_j, d_{j+1}, d_{j+2}, ...\}
\end{equation}


%
%
%
%
%
%

\subsection{$WiSeBE$}

\textit{WiSeBE} is a normalized score dependent of 1) the performance of $C_T$ over $R_W$ and 2) the agreement between all references in $\textbf{R}$.
It is defined as:

\begin{equation}
    \label{eq:wisebe}
    WiSeBE = F1_{R_W} \times R_{G_{AR}} \quad WiSeBE = \left [ 0,1 \right ]\;
\end{equation}

where $F1_{R_W}$ corresponds to the harmonic mean of precision and recall of $C_T$ with respect to $R_W$ (equation \ref{eq:f1}), while $R_{G_{AR}}$ is the agreement ratio defined in~(\ref{eq:AR}).
$R_{G_{AR}}$ can be interpreted as a scaling factor; a low value will penalize the overall \textit{WiSeBE} score given the low agreement between references. By contrast, for a high agreement in $\textbf{R}$ ($R_{G_{AR}} \approx 1$), $WiSeBE \approx F1_{R_W}$.

\begin{equation}
    \label{eq:f1}
    F1_{R_W} = 2 \times  \frac{precision_{R_W} \times recall_{R_W}}{precision_{R_W} + recall_{R_W}}\;
\end{equation}

\begin{equation}
    \label{eq:pr}
    precision_{R_W}=\frac{\sum_{b_j \in C_T}1 \quad[b_j =  1, b_j \in w \quad \forall w \in R_W]}{\sum_{b_j \in C_T}1 \quad[b_j = 1]}\;
\end{equation}

\begin{equation}
    \label{eq:rc}
    recall_{R_W}=\frac{\sum_{w_k \in R_W}1 \quad[w_k \ni b  \quad \forall b \in C_T ]}{p}\;
\end{equation}

\noindent Equations \ref{eq:pr} and \ref{eq:rc} describe precision and recall of $C_T$ with respect to $R_W$. Precision is the number of boundaries $b_j$ inside any window $w_k$ from $R_W$ divided by the total number of boundaries $b_j$ in $C_T$. Recall corresponds to the number of windows $w$ with at least one boundary $b$ divided by the number of windows $w$ in $R_W$.

\section{Evaluating with $WiSeBE$}
\label{sec:exp}

To exemplify the $WiSeBE$ score we evaluated and compared the performance of two different SBD systems over a set of YouTube videos in a multi-reference enviroment.
The first system (S1) employs a Convolutional Neural Network to determine if the middle word of a sliding window corresponds to a SU boundary or not \cite{gonzalez2018transcripts}.
The second approach (S2) by contrast, introduces a bidirectional Recurrent Neural Network model with attention mechanism for boundary detection \cite{tilk2016}.

In a first glance we performed the evaluation of the systems against each one of the references independently. Then, we implemented a multi-reference evaluation with $WiSeBE$.

\subsection{Dataset}

We focused evaluation over a small but diversified dataset composed by 10 YouTube videos in the English language in the news context.
The selected videos cover different topics like technology, human rights, terrorism and politics with a length variation between 2 and 10 minutes.
To encourage the diversity of content format we included newscasts, interviews, reports and round tables.

During the transcription phase we opted for a manual transcription process because we observed that using transcripts from an ASR system will difficult in a large degree the manual segmentation process.
The number of words per transcript oscilate between 271 and 1,602 with a total number of 8,080.

%

We gave clear instructions to three evaluators ($ref_1,ref_2,ref_3$) of how segmentation was needed to be perform, including the SU concept and how punctuation marks were going to be taken into account. 
Periods (.), question marks (?), exclamation marks (!) and semicolons (;) were considered SU delimiters (boundaries) while colons (:) and commas (,) were considered as internal SU marks.
The number of segments per transcript and reference can be seen in Table \ref{tb:manbound}.
An interesting remark is that $ref_3$ assigns about $43\%$ less boundaries than the mean of the other two references. 

\begin{table}
\caption{Manual dataset segmentation}
\label{tb:manbound}
\begin{center}
\begin{tabular}{ c  c  c  c  c  c  c  c c  c  c | c }

\hline
Reference & $v_1$ & $v_2$ & $v_3$ & $v_4$ & $v_5$ & $v_6$ & $v_7$ & $v_8$ & $v_9$ & $v_{10}$ & total \\
\hline
$ref_1$ & 38 & 42 & 17 & 11 & 55 & 87 & 109 & 72 & 55 & 16 & 502   \\

$ref_2$ & 33 &42 & 16 & 14 & 54 & 98 & 92 & 65 & 51 & 20 & 485   \\

$ref_3$ & 23 & 20 & 10 & 6 & 39 & 39 & 76 & 30 & 29 & 9 & 281   \\
\hline
\end{tabular}
\end{center}
\end{table}

\subsection{Evaluation}

We ran both systems (S1 \& S2) over the manually transcribed videos obtaining the number of boundaries shown in Table \ref{tb:autobound}.
In general, it can be seen that S1 predicts $27\%$ more segments than S2.
This difference can affect the performance of S1, increasing its probabilities of false positives.

\begin{table}
\caption{Automatic dataset segmentation}
\label{tb:autobound}
\begin{center}
\begin{tabular}{ c  c  c  c  c  c  c  c  c  c  c | c }
\hline
System & $v_1$ & $v_2$ & $v_3$ & $v_4$ & $v_5$ & $v_6$ & $v_7$ & $v_8$ & $v_9$ & $v_{10}$ & total \\
\hline
S1&53  &38  &15  &13  &54  &108 &106 &70  &71  &11 & 539   \\

S2& 38 &37 &12 &11 &36 &92 &86 &46 &53 &13 & 424   \\
\hline

\end{tabular}
\end{center}
\end{table}

Table \ref{tb:res1} condenses the performance of both systems evaluated against each one of the references independently.
If we focus on F1 scores, performance of both systems varies depending of the reference.
For $ref_1$, S1 was better in 5 occasions with respect of S2; S1 was better in 2 occasions only for $ref_2$; S1 overperformed S2 in 3 occasions concerning $ref_3$ and in 4 occasions for $mean$ (\textbf{bold}).

Also from Table \ref{tb:res1} we can observe that $ref_1$ has a bigger similarity to S1 in 5 occasions compared to other two references, while $ref_2$ is more similar to S2 in 7 transcripts (\underline{underline}).

\begin{table}
\caption{Independent multi-reference evaluation}
\label{tb:res1}
\begin{center}
\begin{tabular}{ c  c  c  c  c  c  c  c  c  c  c | c  c  c }
\hline
&&\multicolumn{3}{c}{$ref_1$} & \multicolumn{3}{c}{$ref_2$}  & \multicolumn{3}{c}{$ref_3$} & \multicolumn{3}{|c}{$mean$}\\ 
\hline
Transcript& System & P & R & F1 & P & R & F1 & P & R & F1 & P & R & F1\\
\hline
\multirow{2}{1em}{$v_1$} & S1 & 0.396 & 0.553 & 0.462 & 0.377 & 0.606 & \underline{0.465} & 0.264 & 0.609 & 0.368 & 0.346 & 0.589 & 0.432 \\ 
& S2 & 0.474 & 0.474 & \textbf{0.474} & 0.474 & 0.545 & \underline{\textbf{0.507}} & 0.368 & 0.6087 & \textbf{0.459} & 0.439 & 0.543 & \textbf{0.480} \\
\hline
\multirow{2}{1em}{$v_2$} & S1 & 0.605 & 0.548 & \textbf{0.575} & 0.711 & 0.643 & \underline{\textbf{0.675}} & 0.368 & 0.700 & \textbf{0.483} & 0.561 & 0.630 & \textbf{0.578} \\
& S2 & 0.595 & 0.524 & 0.557 & 0.676 & 0.595 & \underline{0.633} & 0.351 & 0.650 & 0.456 & 0.541 & 0.590 & 0.549 \\
\hline
\multirow{2}{1em}{$v_3$} & S1 & 0.333 & 0.294 & \underline{0.313} & 0.267 & 0.250 & 0.258 & 0.200 & 0.300 & 0.240 & 0.267 & 0.281 & 0.270 \\ 
& S2 & 0.417 & 0.294 & \textbf{0.345} & 0.417 & 0.313 & \underline{\textbf{0.357}} & 0.250 & 0.300 & \textbf{0.273} & 0.361 & 0.302 & \textbf{0.325} \\
\hline
\multirow{2}{1em}{$v_4$} & S1 & 0.615 & 0.571 & \underline{0.593} & 0.462 & 0.545 & 0.500 & 0.308 & 0.667 & 0.421 & 0.462 & 0.595 & 0.505 \\
& S2 & 0.909 & 0.714 & \textbf{0.800} & 0.818 & 0.818 & \underline{\textbf{0.818}} & 0.455 & 0.833 & \textbf{0.588} & 0.727 & 0.789 & \textbf{0.735} \\
\hline
\multirow{2}{1em}{$v_5$} & S1 & 0.630 & 0.618 & \underline{\textbf{0.624}} & 0.593 & 0.593 & \textbf{0.593} & 0.481 & 0.667 & \textbf{0.560} & 0.568 & 0.626 & \textbf{0.592} \\ 
& S2 & 0.667 & 0.436 & \underline{0.527} & 0.611 & 0.407 & 0.489 & 0.500 & 0.462 & 0.480 & 0.593 & 0.435 & 0.499 \\
\hline
\multirow{2}{1em}{$v_6$} & S1 & 0.491 & 0.541 & \underline{\textbf{0.515}} & 0.454 & 0.563 & 0.503 & 0.213 & 0.590 & 0.313 & 0.386 & 0.565 & 0.443\\
& S2 & 0.500 & 0.469 & 0.484 & 0.522 & 0.552 & \underline{\textbf{0.536}} & 0.250 & 0.590 & \textbf{0.351} & 0.4234 & 0.537 & \textbf{0.457} \\
\hline
\multirow{2}{1em}{$v_7$} & S1 & 0.594 & 0.578 & \underline{\textbf{0.586}} & 0.462 & 0.533 & 0.495 & 0.406 & 0.566 & 0.473 & 0.487 & 0.559 & 0.518 \\ 
& S2 & 0.663 & 0.523 & \underline{0.585} & 0.558 & 0.522 & \textbf{0.539} & 0.465 & 0.526 & \textbf{0.494} & 0.562 & 0.524 & \textbf{0.539} \\
\hline
\multirow{2}{1em}{$v_8$} & S1 & 0.443 & 0.477 & 0.459 & 0.514 & 0.500 & \underline{0.507} & 0.229 & 0.533 & 0.320 & 0.395 & 0.503 & 0.429\\ 
& S2 & 0.609 & 0.431 & \textbf{0.505} & 0.652 & 0.417 & \underline{\textbf{0.508}} & 0.370 & 0.567 & \textbf{0.447} & 0.543 & 0.471 & \textbf{0.487} \\
\hline
\multirow{2}{1em}{$v_9$} & S1 & 0.437 & 0.564 & 0.492 & 0.451 & 0.627 & \underline{0.525} & 0.254 & 0.621 & 0.360 & 0.380 & 0.603 & \textbf{0.459} \\ 
& S2 & 0.623 & 0.600 & \underline{\textbf{0.611}} & 0.585 & 0.608 & \textbf{0.596} & 0.321 & 0.586 & \textbf{0.414} & 0.509 & 0.598 & 0.541 \\
\hline
\multirow{2}{1em}{$v_{10}$} & S1 & 0.818 & 0.450 & \underline{\textbf{0.581}} & 0.818 & 0.450 & \underline{\textbf{0.581}} & 0.455 & 0.556 & \textbf{0.500} & 0.697 & 0.523 & \textbf{0.582} \\ 
& S2 & 0.692 & 0.450 & 0.545 & 0.615 & 0.500 & \underline{0.552} & 0.308 & 0.444 & 0.364 & 0.538 & 0.4645 & 0.487 \\

\hline
\multirow{2}{2em}{mean scores} & S1 & \multicolumn{2}{c}{---} & \underline{0.520} & \multicolumn{2}{c}{---}  & 0.510 & \multicolumn{2}{c}{---}  & 0.404 & \multicolumn{2}{c}{---} & 0.481 \\ 
& S2 & \multicolumn{2}{c}{---} & \textbf{0.543} & \multicolumn{2}{c}{---}  & \textbf{\underline{0.554}} & \multicolumn{2}{c}{---}  & \textbf{0.433} & \multicolumn{2}{c}{---} & \textbf{0.510} \\
\hline
\end{tabular}
\end{center}
\end{table}

After computing the mean F1 scores over the transcripts, it can be concluded that in average S2 had a better performance segmenting the dataset compared to S1, obtaining a F1 score equal to 0.510.
But... What about the complexity of the dataset?
Regardless all references have been considered, nor agreement or disagreement between them has been taken into account.

\vspace{5mm}
\noindent All values related to the $WiSeBE$ score are displayed in Table \ref{tb:wisebeeval}.
The Agreement Ratio ($R_{G_{AR}}$) between references oscillates between 0.525 for $v_8$ and 0.767 for $v_5$.
The lower the $R_{G_{AR}}$, the bigger the penalization $WiSeBE$ will give to the final score.
A good example is S2 for transcript $v_4$ where $F1_{R_W}$ reaches a value of 0.800, but after considering $R_{G_{AR}}$ the $WiSeBE$ score falls to 0.462.

It is feasible to think that if all references are taken into account at the same time during evaluation ($F1_{R_W}$), the score will be bigger compared to an average of independent evaluations ($F1_{mean}$); however this is not always true.
That is the case of S1 in $v10$, which present a slight decrease for $F1_{R_W}$ compared to $F1_{mean}$.

An important remark is the behavior of S1 and S2 concerning $v_6$.
If evaluated without considering any (dis)agreement between references ($F1_{mean}$), S2 overperforms S1; this is inverted once the systems are evaluated with $WiSeBE$.

\begin{table}
\caption{$WiSeBE$ evaluation}
\label{tb:wisebeeval}
\begin{center}
\begin{tabular}{ c  c  c | c  c  c }
\hline
Transcript& System & $F1_{mean}$&$F1_{R_W}$ & $R_{G_{AR}}$ & $WiSeBE$\\
\hline
\multirow{2}{1em}{$v_1$} & S1 & 0.432 & 0.495 & \multirow{2}{2em}{0.691} & 0.342  \\ 
& S2 & \textbf{0.480} & 0.513 &  & \textbf{0.354}  \\
\hline
\multirow{2}{1em}{$v_2$} & S1 & \textbf{0.578} & 0.659 & \multirow{2}{2em}{0.688} & \textbf{0.453}  \\ 
& S2 & 0.549 & 0.595 &  & 0.409  \\
\hline
\multirow{2}{1em}{$v_3$} & S1 & 0.270 & 0.303 & \multirow{2}{2em}{0.684} & 0.207  \\ 
& S2 & \textbf{0.325} & 0.400 &  & \textbf{0.274}  \\
\hline
\multirow{2}{1em}{$v_4$} & S1 & 0.505 & 0.593 & \multirow{2}{2em}{0.578} & 0.342 \\ 
& S2 & \textbf{0.735} & 0.800 &  & \textbf{0.462}  \\
\hline
\multirow{2}{1em}{$v_5$} & S1 & \textbf{0.592} & 0.614 & \multirow{2}{2em}{0.767} & \textbf{0.471}  \\
& S2 & 0.499 & 0.500 &  & 0.383   \\
\hline
\multirow{2}{1em}{$v_6$} & S1 & 0.443 & 0.550 & \multirow{2}{2em}{0.541} & \textbf{0.298}  \\ 
& S2 & \textbf{0.457} & 0.535 &  & 0.289  \\
\hline
\multirow{2}{1em}{$v_7$} & S1 & 0.518 & 0.592 & \multirow{2}{2em}{0.617} & 0.366 \\ 
& S2 & \textbf{0.539} & 0.606 &  & \textbf{0.374}\\
\hline
\multirow{2}{1em}{$v_8$} & S1 & 0.429 & 0.494 & \multirow{2}{2em}{0.525} & 0.259  \\ 
& S2 & \textbf{0.487} & 0.508 &  & \textbf{0.267} \\
\hline
\multirow{2}{1em}{$v_9$} & S1 & 0.459 & 0.569 & \multirow{2}{2em}{0.604} & 0.344  \\ 
& S2 & \textbf{0.541} & 0.667 &  & \textbf{0.403}    \\
\hline
\multirow{2}{1em}{$v_{10}$} & S1 & \textbf{0.582} & 0.581 & \multirow{2}{2em}{0.619} & \textbf{0.359}  \\ 
& S2 & 0.487 & 0.545 &  & 0.338  \\
\hline
\multirow{2}{2em}{mean scores} & S1 & 0.481 & 0.545 & \multirow{2}{2em}{0.631} & 0.344  \\ 
& S2 & \textbf{0.510} & 0.567 &  & \textbf{0.355} \\
\hline
\end{tabular}
\end{center}
\end{table}

\section{Discussion}
\label{sec:disc}

\subsection{$R_{G_{AR}}$ and Fleiss' Kappa correlation}

In Section \ref{sec:WiSeBE} we described the $WiSeBE$ score and how it relies on the $R_{G_{AR}}$ value to scale the performance of $C_T$ over $R_W$.
$R_{G_{AR}}$ can intuitively be consider an agreement value over all elements of $\textbf{R}$.
To test this hypothesis, we computed the Pearson correlation coefficient ($PCC$) \cite{pearson1895note}  between $R_{G_{AR}}$ and the Fleiss' Kappa \cite{fleiss1971measuring} of each video in the dataset ($\kappa _R$). 

\begin{table}
\caption{Aggrement within dataset}
\label{tb:arvskappa}
\begin{center}
\begin{tabular}{ c  c  c  c  c  c  c  c  c  c  c }
\hline
Agreement metric & $v_1$ & $v_2$ & $v_3$ & $v_4$ & $v_5$ & $v_6$ & $v_7$ & $v_8$ & $v_9$ & $v_{10}$ \\
\hline
$R_{G_{AR}}$ &0.691 & 0.688 & 0.684 & 0.578 & 0.767 & 0.541 & 0.617 & 0.525 & 0.604 & 0.619    \\

$\kappa _R$ & 0.776 & 0.697 & 0.757 & 0.696 & 0.839 & 0.630 & 0.743 & 0.655 & 0.704 & 0.718  \\
\hline
\end{tabular}
\end{center}
\end{table}

A linear correlation between $R_{G_{AR}}$ and $\kappa _R$ can be observed in Table \ref{tb:arvskappa}.
This is confirmed by a $PCC$ value equal to $0.890$, which means a very strong positive linear correlation between them.

\subsection{$F1_{mean}$ vs. $WiSeBE$}

Results form Table \ref{tb:wisebeeval} may give an idea that $WiSeBE$ is just an scaled $F1_{mean}$.
While it is true that they show a linear correlation, $WiSeBE$ may produce a different system ranking than $F1_{mean}$ given the integral multi-reference principle it follows.
However, what we consider the most profitable about $WiSeBE$ is the twofold inclusion of all available references it performs.
First, the construction of $R_W$ to provide a more inclusive reference against to whom be evaluated and then, the computation of $R_{G_{AR}}$, which scales the result depending of the agreement between references.

\section{Conclusions}
\label{sec:conc}

In this paper we presented WiSeBE, a semi-automatic multi-reference sentence boundary evaluation protocol based on the necessity of having a more reliable way for evaluating the SBD task.
We showed how $WiSeBE$ is an inclusive metric which not only evaluates the performance of a system against all references, but also takes into account the agreement between them.
According to your point of view, this inclusivity is very important given the difficulties that are present when working with spoken language and the possible disagreements that a task like SBD could provoke.

$WiSeBE$ shows to be correlated with standard SBD metrics, however we want to measure its correlation with extrinsic evaluations techniques like automatic summarization and machine translation.

\section*{Acknowledgments}

We would like to acknowledge the support of CHIST-ERA for funding this work through the Access Multilingual Information opinionS (AMIS), (France - Europe) project.

We also like to acknowledge the support given by the Prof. Hanifa Boucheneb from VERIFORM Laboratory (\'Ecole Polytechnique de Montr\'eal).

\bibliographystyle{splncs04}
\bibliography{paper}

\begin{thebibliography}{10}
\providecommand{\url}[1]{\texttt{#1}}
\providecommand{\urlprefix}{URL }
\providecommand{\doi}[1]{https://doi.org/#1}

\bibitem{bohac2012post}
Bohac, M., Blavka, K., Kucharova, M., Skodova, S.: Post-processing of the
  recognized speech for web presentation of large audio archive. In:
  Telecommunications and Signal Processing (TSP), 2012 35th International
  Conference on. pp. 441--445. IEEE (2012)

\bibitem{brum2016sentiment}
Brum, H., Araujo, F., Kepler, F.: Sentiment analysis for brazilian portuguese
  over a skewed class corpora. In: International Conference on Computational
  Processing of the Portuguese Language. pp. 134--138. Springer (2016)

\bibitem{che2016punctuation}
Che, X., Wang, C., Yang, H., Meinel, C.: Punctuation prediction for unsegmented
  transcript based on word vector. In: LREC (2016)

\bibitem{fleiss1971measuring}
Fleiss, J.L.: Measuring nominal scale agreement among many raters.
  Psychological bulletin  \textbf{76}(5), ~378 (1971)

\bibitem{fohr2017new}
Fohr, D., Mella, O., Illina, I.: New paradigm in speech recognition: Deep
  neural networks. In: IEEE International Conference on Information Systems and
  Economic Intelligence (2017)

\bibitem{gonzalez2018transcripts}
Gonz{\'a}lez-Gallardo, C.E., Hajjem, M., SanJuan, E., Torres-Moreno, J.M.:
  Transcripts informativeness study: An approach based on automatic
  summarization. In: {Conf{\'e}rence en Recherche d'Information et Applications
  (CORIA)}. Rennes, France (May 2018)

\bibitem{gonzalez2018sentence}
Gonz{\'a}lez-Gallardo, C.E., Torres-Moreno, J.M.: Sentence boundary detection
  for french with subword-level information vectors and convolutional neural
  networks. arXiv preprint arXiv:1802.04559  (2018)

\bibitem{gotoh2000sentence}
Gotoh, Y., Renals, S.: Sentence boundary detection in broadcast speech
  transcripts. In: ASR2000-Automatic Speech Recognition: Challenges for the new
  Millenium ISCA Tutorial and Research Workshop (ITRW) (2000)

\bibitem{hinton2012deep}
Hinton, G., Deng, L., Yu, D., Dahl, G.E., Mohamed, A.r., Jaitly, N., Senior,
  A., Vanhoucke, V., Nguyen, P., Sainath, T.N., et~al.: Deep neural networks
  for acoustic modeling in speech recognition: The shared views of four
  research groups. IEEE Signal Processing Magazine  \textbf{29}(6),  82--97
  (2012)

\bibitem{jamil2015sentence}
Jamil, N., Ramli, M.I., Seman, N.: Sentence boundary detection without speech
  recognition: A case of an under-resourced language. Journal of Electrical
  Systems  \textbf{11}(3) (2015)

\bibitem{kiss2006unsupervised}
Kiss, T., Strunk, J.: Unsupervised multilingual sentence boundary detection.
  Computational Linguistics  \textbf{32}(4),  485--525 (2006)

\bibitem{klejch2016punctuated}
Klejch, O., Bell, P., Renals, S.: Punctuated transcription of multi-genre
  broadcasts using acoustic and lexical approaches. In: Spoken Language
  Technology Workshop (SLT), 2016 IEEE. pp. 433--440. IEEE (2016)

\bibitem{kolavr2012development}
Kol{\'a}{\v{r}}, J., Lamel, L.: Development and evaluation of automatic
  punctuation for french and english speech-to-text. In: Thirteenth Annual
  Conference of the International Speech Communication Association (2012)

\bibitem{kolavr2004automatic}
Kol{\'a}{\v{r}}, J., {\v{S}}vec, J., Psutka, J.: Automatic punctuation
  annotation in czech broadcast news speech. SPECOM{\'{}} 2004  (2004)

\bibitem{liu2006study}
Liu, Y., Chawla, N.V., Harper, M.P., Shriberg, E., Stolcke, A.: A study in
  machine learning from imbalanced data for sentence boundary detection in
  speech. Computer Speech \& Language  \textbf{20}(4),  468--494 (2006)

\bibitem{lu2010better}
Lu, W., Ng, H.T.: Better punctuation prediction with dynamic conditional random
  fields. In: Proceedings of the 2010 conference on empirical methods in
  natural language processing. pp. 177--186. Association for Computational
  Linguistics (2010)

\bibitem{meteer1996modeling}
Meteer, M., Iyer, R.: Modeling conversational speech for speech recognition.
  In: Conference on Empirical Methods in Natural Language Processing (1996)

\bibitem{mrozinski2006automatic}
Mrozinski, J., Whittaker, E.W., Chatain, P., Furui, S.: Automatic sentence
  segmentation of speech for automatic summarization. In: 2006 IEEE
  International Conference on Acoustics Speech and Signal Processing
  Proceedings. vol.~1, pp.~I--I. IEEE (2006)

\bibitem{nicola2013improved}
Nicola, U., Maximilian, B., Paul, V.: Improved models for automatic punctuation
  prediction for spoken and written text. In: Proceedings of INTERSPEECH (2013)

\bibitem{palmer1994adaptive}
Palmer, D.D., Hearst, M.A.: Adaptive sentence boundary disambiguation. In:
  Proceedings of the Fourth Conference on Applied Natural Language Processing.
  pp. 78--83. ANLC '94, Association for Computational Linguistics, Stroudsburg,
  PA, USA (1994)

\bibitem{palmer1997adaptive}
Palmer, D.D., Hearst, M.A.: Adaptive multilingual sentence boundary
  disambiguation. Comput. Linguist.  \textbf{23}(2),  241--267 (Jun 1997)

\bibitem{pearson1895note}
Pearson, K.: Note on regression and inheritance in the case of two parents.
  Proceedings of the Royal Society of London  \textbf{58},  240--242 (1895)

\bibitem{peitz2014better}
Peitz, S., Freitag, M., Ney, H.: Better punctuation prediction with
  hierarchical phrase-based translation. In: Proc. of the Int. Workshop on
  Spoken Language Translation (IWSLT), South Lake Tahoe, CA, USA (2014)

\bibitem{rott2016speech}
Rott, M., {\v{C}}erva, P.: Speech-to-text summarization using automatic phrase
  extraction from recognized text. In: International Conference on Text,
  Speech, and Dialogue. pp. 101--108. Springer (2016)

\bibitem{shriberg1996word}
Shriberg, E., Stolcke, A.: Word predictability after hesitations: a
  corpus-based study. In: Spoken Language, 1996. ICSLP 96. Proceedings., Fourth
  International Conference on. vol.~3, pp. 1868--1871. IEEE (1996)

\bibitem{stevenson2000experiments}
Stevenson, M., Gaizauskas, R.: Experiments on sentence boundary detection. In:
  Proceedings of the sixth conference on Applied natural language processing.
  pp. 84--89. Association for Computational Linguistics (2000)

\bibitem{stolcke1996automatic}
Stolcke, A., Shriberg, E.: Automatic linguistic segmentation of conversational
  speech. In: Spoken Language, 1996. ICSLP 96. Proceedings., Fourth
  International Conference on. vol.~2, pp. 1005--1008. IEEE (1996)

\bibitem{strassel2003simple}
Strassel, S.: Simple metadata annotation specification v5. 0, linguistic data
  consortium.
  http://www.ldc.upenn.edu/projects/MDE/Guidelines/SimpleMDE\_V5.0.pdf  (2003)

\bibitem{tilk2016}
Tilk, O., Alum{\"a}e, T.: Bidirectional recurrent neural network with attention
  mechanism for punctuation restoration. In: Interspeech 2016 (2016)

\bibitem{treviso2017evaluating}
Treviso, M.V., Shulby, C.D., Aluisio, S.M.: Evaluating word embeddings for
  sentence boundary detection in speech transcripts. arXiv preprint
  arXiv:1708.04704  (2017)

\bibitem{ueffing2013improved}
Ueffing, N., Bisani, M., Vozila, P.: Improved models for automatic punctuation
  prediction for spoken and written text. In: Interspeech. pp. 3097--3101
  (2013)

\bibitem{wang2010automatic}
Wang, W., Tur, G., Zheng, J., Ayan, N.F.: Automatic disfluency removal for
  improving spoken language translation. In: Acoustics speech and signal
  processing (icassp), 2010 ieee international conference on. pp. 5214--5217.
  IEEE (2010)

\bibitem{xu2014deep}
Xu, C., Xie, L., Huang, G., Xiao, X., Chng, E.S., Li, H.: A deep neural network
  approach for sentence boundary detection in broadcast news. In: Fifteenth
  annual conference of the international speech communication association
  (2014)

\bibitem{yu2016automatic}
Yu, D., Deng, L.: Automatic speech recognition. Springer (2016)

\end{thebibliography}
\end{document}